\documentclass[final]{cvpr}

\usepackage{times}
\usepackage{epsfig}
\usepackage{graphicx}
\usepackage{amsmath}
\usepackage{amssymb}
\usepackage{multirow}
\usepackage{color, colortbl}
\usepackage{amsmath,amsfonts,amsthm,bm}

\usepackage{booktabs}
\usepackage{mathtools}
\usepackage{multirow}
\usepackage{makecell}

\definecolor{Gray}{gray}{0.85}

\usepackage[pagebackref=true,breaklinks=true,colorlinks,bookmarks=false]{hyperref}

\def\cvprPaperID{2367} 
\def\confYear{CVPR 2021}
\makeatletter
\newcommand{\printfnsymbol}[1]{%
  \textsuperscript{\@fnsymbol{#1}}%
}

\begin{document}
\title{GOSS: Towards Generalized Open-set Semantic Segmentation}

\author{Jie Hong$^{1,2}$, Weihao Li$^{2}$, Junlin Han$^{1,2}$, Jiyang Zheng$^{1,2}$, Pengfei Fang$^{1,2}$, 
\and
Mehrtash Harandi$^{3}$, Lars Petersson$^{2}$ \\
$^{1}$Australian National University, $^{2}$Data61-CSIRO, $^{3}$Monash University \\
{\tt\small \{jie.hong,junlin.han,jiyang.zheng,pengfei.fang\}@anu.edu.au},\\
{\tt\small \{weihao.li,lars.petersson\}@data61.csiro.au},
{\tt\small mehrtash.harandi@monash.edu},
}
\maketitle
\thispagestyle{empty}
\pagestyle{empty}

\begin{abstract}
In this paper, we present and study a new image segmentation task, called \textit{Generalized Open-set Semantic Segmentation (GOSS)}. Previously, with the well-known open-set semantic segmentation (OSS), the intelligent agent only detects the unknown regions without further processing, limiting their perception of the environment. It stands to reason that a further analysis of the detected unknown pixels would be beneficial. Therefore, we propose GOSS, which unifies the abilities of two well-defined segmentation tasks, \ie, OSS and generic segmentation (GS), in a holistic way. Specifically, GOSS classifies pixels as belonging to known classes, and clusters (or groups)\footnote{In the remainder of the paper, we interchangeably use the `cluster' and `group'.} of pixels of unknown class are labelled as such. To evaluate this new expanded task, we further propose a metric which balances the pixel classification and clustering aspects. Moreover, we build benchmark tests on top of existing datasets and propose a simple neural architecture as a baseline, which jointly predicts pixel classification and clustering under open-set settings. Our experiments on multiple benchmarks demonstrate the effectiveness of our baseline. We believe our new GOSS task can produce an expressive image understanding for future research. Code will be made available.
\end{abstract}

\section{Introduction}
In the deep learning era, image segmentation, especially class-specific semantic segmentation (SS), has received significant progress \cite{long2015fully,chen2017deeplab,chen2017rethinking,chen2018encoder,cheng2021per,zheng2021rethinking}. The goal of the SS task is to predict the class label of each pixel in an image from a set of predefined object classes. Adjacent pixels naturally belong together to form a segment when they share the same object category. Despite the considerable improvement, most SS settings follow a strong closed-set assumption that training and test data come from the same set of \textit{known} object classes \cite{shotton2006textonboost,cordts2016cityscapes,zhou2017scene,caesar2018coco}. However, this assumption is rarely the case in practice. It limits the generalization of segmentation models to \textit{unknown} classes which models do not see during training. 

\begin{figure}[t]
\centering
	\includegraphics[width=1.0\linewidth]{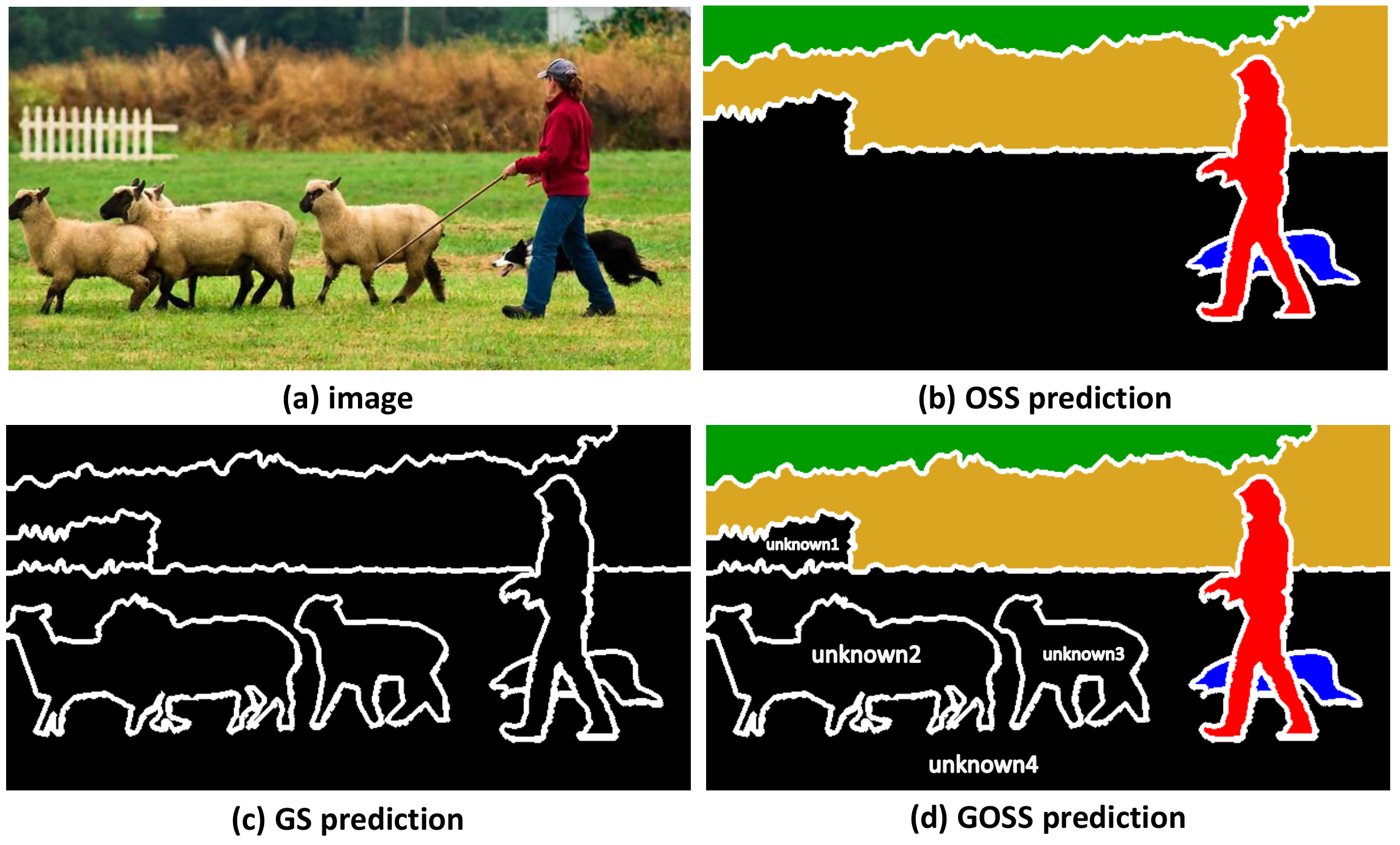}
 	\caption{Different tasks of image segmentation. For a given input image \textbf{(a)} that contains both known (`person', `dog' and `vegetation') and unknown objects (`sheep', `rail' and `grass'), we show: \textbf{(b)} open-set semantic segmentation (OSS) by pixel identification, \textbf{(c)} generic segmentation (GS) by pixel clustering,  and \textbf{(d)} generalized open-set semantic segmentation (GOSS).}
 	\label{figure_wholistic}
 	\vspace{-2mm}
\end{figure}

Recently, open-set semantic segmentation (OSS) \cite{hendrycks2019scaling,lis2019detecting,Cen_2021_ICCV,chan2021segmentmeifyoucan} has been proposed to relax the above assumption, which aims to segment an image that contains both known and unknown object classes. Different from SS, OSS identifies the unknown region where pixels belong to unknown classes.
Although OSS aims to identify pixels that do not belong to one of the known classes, it does not provide any further segmentation amongst those identified unknown pixels. We argue that such a setting of OSS may limit the broad usage of vision-based intelligent agents when they encounter unfamiliar scenes where several unknown object classes are adjacent to each other rather than separated. Consider a scenario that an intelligent agent goes into a new scene like shown in Figure \ref{figure_wholistic} (a). OSS leaves the whole unknown region as a large segment without further processing (see `black region' in Figure \ref{figure_wholistic} (b)). Such insufficient information, provided by the OSS setting, will affect the decision-making for the intelligent agents. Hence, we raise fundamental questions: how to improve OSS for generating richer representations for images in real-world scenes? Moreover, what is a more expressive and versatile image segmentation task beyond OSS? In real-world applications, these questions are crucial. Again considering the example in Figure \ref{figure_wholistic} (a), self-driving cars may be assisted to avoid potential obstacles if `unknown sheep' or `unknown rail' inside `unknown grass' can be found in the OSS prediction (see Figure (b)). Another possible practical example is that new detected `objects' inside unknown regions of images could help accelerate the process of data annotation, especially when images from unfamiliar scenes are being labeled.

Towards the goal of better handling unknown regions in an image, inspired by the perception system from humans that they can jointly recognize the previously known objects and easily group unknown areas into different segments even though they do not know the categories of those unknown objects, this paper studies a new type of semantic image segmentation task called \textit{Generalized Open-set Semantic Segmentation (GOSS)}. It aims to classify pixels belonging to known classes and group the unknown pixels (see Figure \ref{figure_wholistic} (d) for GOSS prediction). As we can see from Figure \ref{figure_wholistic} (d), `unknown rail' (or `unknown 1') and `unknown sheep' (or `unknown2') are segmented out from `unknown grass' (or `unknown4'). GOSS takes the advantage of generic segmentation (GS) which groups pixels into segments sharing similarities \cite{shi2000normalized,comaniciu2002mean,felzenszwalb2004efficient,arbelaez2010contour,arbelaez2010contour,isaacs2020enhancing,wan2020super}. Compared to OSS, GOSS is able to detect more `objects' inside unknown regions.

To enable intelligent agents to perform GOSS, we first build benchmarks using existing segmentation datasets, \ie, COCO-Stuff \cite{caesar2018coco} and Cityscapes \cite{cordts2016cityscapes}. We split the full set of object categories into two sets: known classes and unknown classes. We keep the semantic annotations of known categories. For unknown categories, we use connectivity labeling \cite{wan2020super} to convert their original semantic annotations into clustering ground truths. Along with the available datasets, it also requires a metric that can encompass the quality of both OSS and GS. Although there are many existing metrics for segmentation tasks, such metrics are limited to measuring a single setting. In this work, we introduce a metric, termed \underline{G}OSS \underline{Q}uality ($\mathrm{GQ}$), which evaluates the segmentation quality of both known and unknown objects. Having the datasets and evaluation metrics at hand, we further establish an end-to-end trainable framework, namely, \underline{G}OSS \underline{S}egme\underline{T}or (GST). The proposed GST adopts a dual-branch architecture with a shared backbone network. To perform the GOSS task, one branch conducts pixel classification, and the other performs pixel clustering. Moreover, to learn more discriminative embeddings and thus better process unknown objects, we leverage the pixel-wise contrastive learning loss into the training.

In summary, our contributions are as follows: 1) We present a new image segmentation task called GOSS, which jointly classifies known pixels and groups identified unknown pixels from OSS; 2) We propose a metric that measures the quality of both pixel classification and pixel clustering, under open-set settings; 3) According to settings of GOSS, we build benchmarks by customizing existing datasets; 4) We show a simple yet effective model, GST, to facilitate future research. 
\section{Related Work}
In computer vision, image segmentation is one of the most widely explored problems. During the history of image segmentation research, novel segmentation tasks have played a crucial role in driving research directions and innovations. We provide comparisons between our new setting and relevant older tasks in Table \ref{tab:tasks}. 

\begin{table}
  \centering
  \resizebox{0.475\textwidth}{!}{
  \begin{tabular}{@{}lll@{}}
    \hline
    \textbf{Task} & \textbf{Known classes} & \textbf{Unknown classes}
    \\
    \hline
    \textbf{Generic Segmentation}   &Cluster      & Cluster  \\
    \textbf{Open-set Semantic Segmentation}  &Classify     &Identify \\
    \hline
    \textbf{Generalized Open-set Semantic Segmentation} &Classify     &Identify $\&$ Cluster \\
    \hline
  \end{tabular}}
  \vspace{0.5mm}
  \caption{Comparisons of different image segmentation tasks. Compared to traditional segmentation tasks, GOSS takes better care of unknown objects.  
  } 
  \label{tab:tasks}
  \vspace{-2mm}
\end{table}

\noindent \textbf{Open-set Semantic Segmentation (OSS)}. OSS, which is capable of identifying unknown objects, has been developed significantly in recent years. 
Performing OSS is essential for intelligent agents as they work in open-set settings where many objects are never seen before. 
A natural solution, studied in \cite{hendrycks2019scaling,jung2021standardized,chan2021entropy}, determines the unknown regions by directly computing the anomaly score from logit or confidence vectors provided by the model classifier. Alternatively, synthesis approaches \cite{lis2019detecting,xia2020synthesize,di2021pixel,vojir2021road,kong2021opengan} are proposed to detect unexpected anomalies from reconstructed images. 
In addition, the work \cite{Cen_2021_ICCV} employs metric learning to learn more discriminative features and incrementally label novel classes using a human-in-the-loop approach.
Beyond the existing OSS setting, the proposed GOSS performs the holistic segmentation via classifying the known objects and clustering the unknown objects, such that it provides more expressive information of the environment than OSS. Having richer information at hand, GOSS can benefit practical usages in real-world scenarios.

\noindent \textbf{Generic Segmentation (GS).} 
The task of GS is to find groups of pixels that `go together' \cite{szeliski2010computer}. In the early days of computer vision, the term `image segmentation' and the bottom-up general (non-semantic) segmentation share the same meaning. Recently, it is often termed `generic segmentation' \cite{arbelaez2010contour,isaacs2020enhancing,wan2020super} to distinguish it from other segmentation tasks. 
The pipeline of early segmentation methods consists of first extracting local pixel features such as brightness, color or texture and then clustering these features based on, \eg, mean shift \cite{comaniciu2002mean}, normalized cuts \cite{shi2000normalized}, random walks \cite{grady2006random}, graph-based representations \cite{felzenszwalb2004efficient}, or oriented watershed transform \cite{arbelaez2010contour}.
Learning-based image segmentation methods have now also become popular. DEL \cite{liu2018deep} learns a feature embedding that corresponds to a similarity measure between two adjacent superpixels. Saacs et al. \cite{isaacs2020enhancing} propose pixel-wise representations that reflect how segments are related. Super-BPD \cite{wan2020super} learns super boundary-to-pixel direction to provide a direction similarity between adjacent pixels. Comparing the performance of different image segmentation algorithms, public datasets such as BSDB \cite{martin2001database,arbelaez2010contour} provide human-labeled class-agnostic ground truth. However, they do not provide any semantic information. 

\noindent \textbf{Image Segmentation as a subtask.} Image segmentation is often taken as a subtask jointly solved with other vision problems in a single framework \cite{eigen2015predicting,jafari2017analyzing,kokkinos2017ubernet}, \cite{bleyer2011object,yamaguchi2014efficient,sun2012layered,sevilla2016optical,hane2013joint,hu2019segmentation}. 
Recently, panoptic segmentation \cite{kirillov2019panoptica,kirillov2019panopticb} has become a common image segmentation task by unifying semantic segmentation and instance segmentation.

\section{Format and Metric}
\subsection{Task Format}
Here, the task format for GOSS is formulated at a pixel level. For the $i$th pixel of an image, the GOSS output is defined as a pair $\textbf{goss}_i = (s_i, g_i)$, where the classification label $s_i$ indicates the pixel's semantic class and the clustering (or grouping) label $g_i$ represents the cluster id. Suppose that there are ${N}$ known semantic classes $\textbf{L}^{kn} \in \mathbb{R}^N$ and an unknown class indicator $L^{uk} \in \mathbb{R}$, we have the semantic label set $\textbf{L} = \{\textbf{L}^{kn}, L^{uk}\}$  which is encoded by $\textbf{L}:= \{ 0, ..., N-1, N\}$. In our formulation, each pixel can be predicted as either one of known classes or the unknown class.  In the first case, each pixel must have a semantic label, while the cluster id is not necessary. This is due to the fact that once the $i$th pixel is labelled with $s_i \in \textbf{L}^{kn}$, its corresponding cluster id $g_i$ is invalid (which is denoted by $void$). When the pixel is predicted as the unknown class, then it can be clustered to $g_i$. Hence, the $i$th pixel with known classes (or unknown classes) can be assigned with $\textbf{goss}_i=(s_i, void)$ (or $\textbf{goss}_i=(N, g_i)$). In practice, $s_i$ can be predicted by a classification model, and $g_i$ can be determined after the unknown pixels are clustered.

\subsection{Evaluation Metrics}
\label{section:metrics}
Appropriate evaluation metrics play a fundamental role in driving the popularization of a new image segmentation task \cite{kirillov2019panoptica,cheng2021boundary,everingham2010pascal}. In this subsection, we first briefly review some popular existing metrics for relevant segmentation tasks and then introduce a new metric, tailored for the proposed GOSS. 

\noindent \textbf{Previous Metrics.} Standard metrics for OSS include the false positive rate at 95\% true positive rate (FPR at 95\% TPR), the area under receiver operating characteristics (AUROC) \cite{davis2006relationship,fawcett2006introduction} and the area under the precision-recall (AUPR) \cite{manning1999foundations,saito2015precision}. Such metrics assess the performance based on the overlap of anomaly score distributions between the known and unknown class. However, they are not suited for evaluating GOSS since they do not need to clearly classify the input as a known or unknown class. GOSS requires each input pixel to be explicitly classified as belonging to a known or unknown class, since GS (or clustering) is only performed on the identified unknown pixels. Well-known metrics for GS include the variation of information \cite{meila2005comparing}, probabilistic rand index \cite{rand1971objective}, F-measure \cite{martin2004learning}, and segmentation covering \cite{arbelaez2010contour}. These metrics are initially proposed to evaluate the quality of either data clustering or edge detection. As no multi-class semantic labels are considered, they cannot be directly used to measure the performance of joint GS and OSS.

\noindent \textbf{GOSS Quality.} We borrow the idea of the segment matching from the panoptic quality \cite{kirillov2019panoptica} in panoptic segmentation (PS) and adapt the panoptic quality as GOSS quality to be suitable for evaluating GOSS. As shown in Figure \ref{figure_wholistic} (d), the GOSS output can be viewed as a set of predicted segments (\ie, each segment or cluster has a unique id $\textbf{goss}_i$), which is similar to the panoptic output. However, unlike GOSS, PS is not able to deal with unknown objects. In comparison with PS, GOSS does not differentiate `instance-level' segments for both known and unknown objects. For example, in Figure \ref{figure_wholistic} (d), GOSS does not separate `two sheep' in `unknown2' into instances.

\begin{figure}
\centering
	\includegraphics[width=0.95\linewidth]{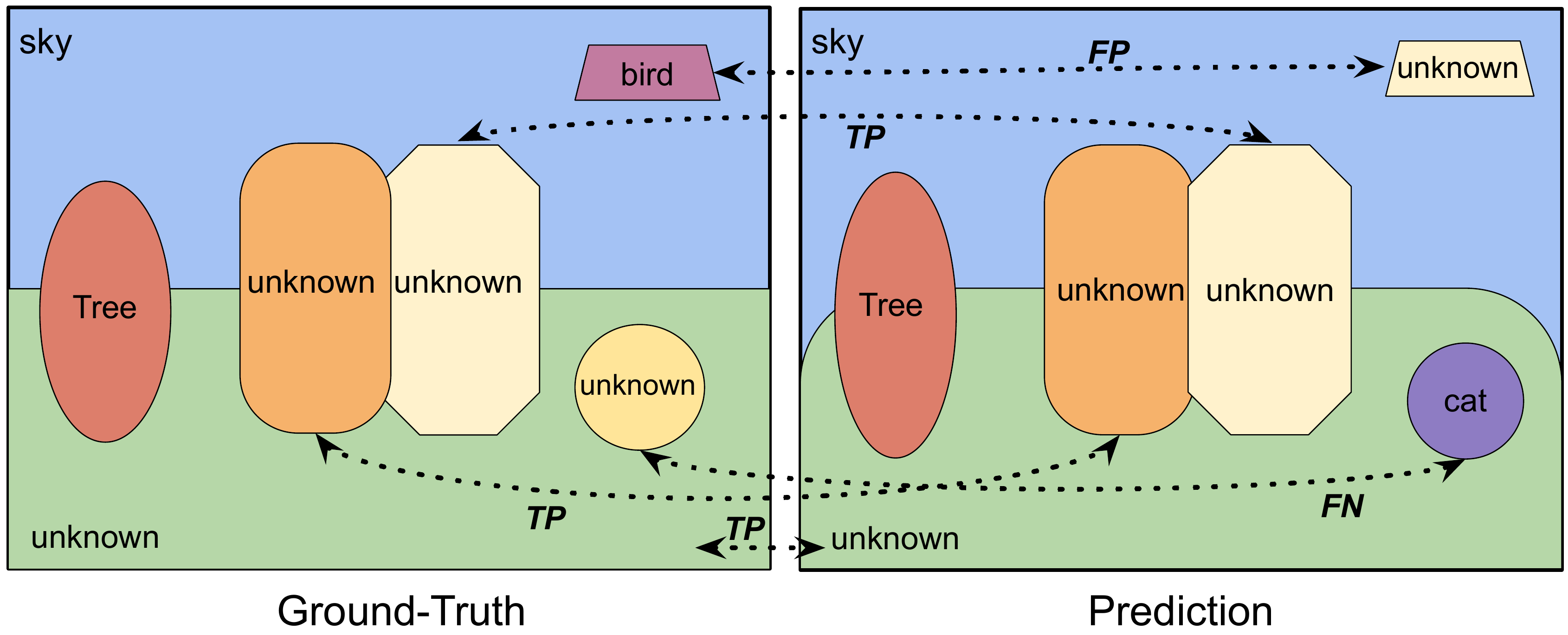}
 	\caption{Toy model of ground truth and predicted GOSS of an image. The predicted segments for `unknown' are partitioned into true positives $\mathrm{TP}^{uk}$, false negatives $\mathrm{FP}^{uk}$, and false positives $\mathrm{FN}^{uk}$.}
 	\label{figure_wq}
 	\vspace{-2mm}
\end{figure}

We treat the unknown pixels as a new class. Thus, there is a total of ${N+1}$ classes of segmentation. As introduced in Figure \ref{figure_wq}, with the segment matching, each predicted segment from GOSS is matched with at most one ground truth segment when their $\mathrm{IoU}$ is higher than 0.5. Pixels belonging to the same segment have the identical $\textbf{goss}$. We let $\mathrm{GQ}^{kn}$ be the average GOSS quality over ${N}$ known classes. Accordingly, $\mathrm{GQ}^{uk}$ is the GOSS quality of the unknown class:
\begin{equation}
\begin{aligned}
\mathrm{GQ}^{kn} &= \frac{1}{N} \sum_{k \in \textbf{L}^{kn}} \frac{\sum_{(u, \hat{u}) \in \mathrm{TP}_{k}^{kn}} \mathrm{IoU}(u, \hat{u})}{|\mathrm{TP}_{k}^{kn}| + \frac{1}{2}|\mathrm{FP}_{k}^{kn}| + \frac{1}{2}|\mathrm{FN}_{k}^{kn}|} \\
\mathrm{GQ}^{uk} &= \frac{\sum_{(u, \hat{u}) \in \mathrm{TP}^{uk}} \mathrm{IoU}(u, \hat{u})}{|\mathrm{TP}^{uk}| + \frac{1}{2}|\mathrm{FP}^{uk}| + \frac{1}{2}|\mathrm{FN}^{uk}|}
\end{aligned}\label{metric}
\end{equation}
where $\mathrm{IoU}(u, \hat{u})$ calculates the Intersection over Union value for the predicted segment $u$ and the ground-truth segment $\hat{u}$.
Furthermore, $\mathrm{TP}_{k}^{kn}$, $\mathrm{FP}_{k}^{kn}$ and $\mathrm{FN}_{k}^{kn}$ denote true positives, false positives and false negatives for the $k$th known class, respectively. Similarly, $\mathrm{GQ}^{uk}$ is obtained specially for the unknown class with its true positives $\mathrm{TP}^{uk}$, false positives $\mathrm{FP}^{uk}$ and false negatives $\mathrm{FN}^{uk}$. 

The metrics $\mathrm{GQ}^{kn}$ and $\mathrm{GQ}^{uk}$ are computed based on the GOSS output (see Figure \ref{figure_wholistic} (d)). The known and unknown segments on the GOSS prediction are evaluated separately via $\mathrm{GQ}^{kn}$ and $\mathrm{GQ}^{uk}$. A unified metric is required to simplify the evaluation. Thus, we define a new metric GOSS Quality ($\mathrm{GQ}$) as: 
\begin{equation}
\mathrm{GQ}= \lambda \cdot \mathrm{GQ}^{kn} + (1-\lambda) \cdot \mathrm{GQ}^{uk}
\end{equation}
where we set $\lambda$ as the most natural number, 0.5, throughout the paper. The reason behind this design is that if we simply average $\mathrm{GQ}$ over $N+1$ classes, then the ratio between known and unknown is significantly biased ($N:1$). $\mathrm{GQ}$ takes care of the known and unknown segments equally. We also introduce $\mathrm{GQ}^{clu}$, the GOSS quality of the pixel clustering (see Figure \ref{figure_wholistic}(b)) before fusing it with the pixel classification\&identification. Refer to the supplementary material for more details of $\mathrm{GQ}^{clu}$. 

\subsection{Challenges}
We argue that, compared with the existing OSS task, GOSS poses more challenges since it requires not only identifying unknown regions in an image but also grouping pixels within these regions into clusters. Clustering objects from unknown classes greatly increases the task difficulty. 

\section{Methodology}
\label{section:method}
\begin{figure*}
\centering
	\includegraphics[width=1.0\linewidth]{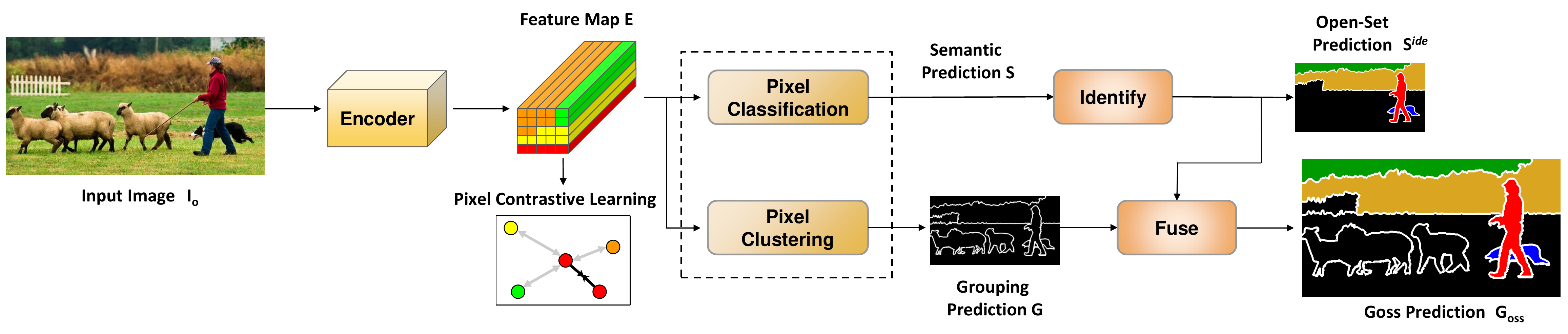}
 	\caption{\textbf{The framework of GOSS Segmentor (GST).} The input image is fed into the encoder for feature extraction. The dual-branch heads are jointly trained for the pixel classification and clustering. Furthermore, the pixel-wise contrastive learning is leveraged to learn discriminative feature embeddings. The pixel identification module is designed to recognize sets of pixels of the unknown class from the semantic prediction. The final GOSS output is generated by fusing the identified semantic prediction and the grouping prediction.}
 	\label{figure_baseline}
\end{figure*}

In order to effectively perform GOSS, we propose a baseline framework. The baseline is mainly comprised of five components: the shared encoder, the pixel classification branch, the pixel clustering branch, the identification module, and the fusion module. Then, we extend the baseline into a more advanced one, GOSS SegmenTor (GST), as shown in Figure \ref{figure_baseline}. More details of our design will be described next.

\subsection{Baseline}
GOSS can be modeled as a unified segmentation task that incorporates the pixel classification and clustering in an open-setting scenario.
Given an image $I_o \in \mathbb{R}^{3 \times \hbar_o \times \omega_o}$, we expect the proposed baseline to simultaneously generate the semantic and grouping predictions. 
Hence, we adopt a dual-branch architecture, with one branch for pixel classification and another for pixel clustering.
As shown in Figure \ref{figure_baseline}, two branches share the same encoder as a feature extractor.
The branch for pixel classification computes a prediction map $\textbf{S} \in \mathbb{R}^{\hbar_o \times \omega_o}$ while the pixel clustering branch outputs a mask map $\textbf{G} \in \mathbb{R}^{\hbar_o \times \omega_o}$ which includes the grouped class-agnostic segments. 
The unknown regions in $\textbf{S}$ are identified, denoted by $\textbf{S}^{ide}$, which is further fused with $\textbf{G}$, obtaining the final GOSS output $\textbf{G}_{oss} \in \mathbb{R}^{2 \times \hbar_o \times \omega_o}$.
The baseline model is jointly trained with two losses: the classification loss $\ell_{cla}$, and the clustering loss $\ell_{clu}$. The total loss is $\ell_{ws} = \alpha_{cla}\ell_{cla} + \alpha_{clu}\ell_{clu}$ where $\alpha_{cla}$ and $\alpha_{clu}$ are positive adjustment weights.

\subsubsection{Pixel Classification}
We train the branch for pixel classification to classify each pixel as one of ${N}$ classes where ${N}$ is the number of predefined known classes. DeepLabV3+ \cite{chen2018encoder}, an existing powerful baseline for semantic segmentation, is chosen as the basic architecture for this branch. The branch is updated under $\ell_{cla}$ which is the cross-entropy loss between the predicted semantic map $\textbf{S}$ and its ground-truth map. 
DeepLabV3+ leverages an encoder-decoder architecture that takes a bottom-up pathway network with features at multiple spatial resolutions, and appends a top-down pathway with lateral connections. The top-down pathway progressively upsamples features starting from the deepest layer of the network while concatenating or adding them with higher-resolution features from the bottom-up pathway. The Atrous Spatial Pyramid Pooling (ASPP) layer \cite{chen2017rethinking} is employed in the DeepLabV3+ model to enlarge the receptive field.

\subsubsection{Pixel Identification}
Pixel identification from OSS is executed to identify sets of pixels of unknown classes from the semantic prediction. Hereafter, we study several recipes for pixel identification with which the identified semantic prediction $\textbf{S}^{ide} \in \mathbb{R}^{\hbar_o \times \omega_o}$ is computed after processing $\textbf{S}$. Common metrics like AUROC and AUPR, of OSS assess the distribution overlap between known and unknown classes, but pixel identification is required to clearly state if the input pixel is known or not. In other words, the binary classification is necessary.

\noindent \textbf{N-model.} In the pixel classification branch it is natural to design the semantic segmentation model with an $N$-dimensional confidence output $\textbf{C} \in \mathbb{R}^{N \times \hbar_o \times \omega_o}$. The N-model is restricted to recognizing the set of predetermined known classes. When an unknown region comes up in a test image, it would be erroneously classified as one of the known classes. To identify unknown pixels based on outputs from the N-model, we employ Maximum Softmax Probability (MSP) \cite{hendrycks2016baseline}, Maximum Unnormalized Logit (MaxLogit) \cite{hendrycks2019scaling} and Deep Metric Learning (DML) \cite{Cen_2021_ICCV}, respectively. Thresholds are used to clearly classify pixels as belonging to a known or unknown class. More details can be found in the supple.

\noindent \textbf{N+1-model.} As opposed to the N-model, the N+1-model \cite{devries2018learning,corbiere2019addressing} contains the unknown class in the output $\textbf{C} \in \mathbb{R}^{(N +1) \times \hbar_o \times \omega_o}$ such that the N+1-model can directly identify the unknown pixels. During the training stage, the N+1-model explicitly takes the 'unlabeled' pixels (\ie, the 'void' pixels) as the unknown pixels. N+1-model is not valid if no `void' pixels are provided.

\subsubsection{Pixel Clustering}
The pixel clustering branch is built in parallel with the pixel classification branch. The goal of this branch is to partition the whole image into clusters. During training, to generate the corresponding annotations, we convert the semantic labeling of the known classes into connectivity labeling by ignoring the previous semantics of each segment. The top-performing method, super boundary-to-pixel direction (Super-BPD) \cite{wan2020super} is selected to establish the branch. The branch is trained in an end-to-end supervised manner as well. Super-BPD applies the ground-truth annotation generated by the distance transform algorithm. Using the Super-BPD model, the representation of boundary-to-pixel direction (BPD) for each pixel is learned ($\ell_{clu} = \ell_{bpd}$). Super-BPDs are extracted based on the initial BPDs using the component-tree computation, followed by graph partitioning to merge super-BPDs into new segments.

\subsubsection{Fusion Module}
The identification module outputs the identified semantic prediction $\textbf{S}^{ide}$. Based on $\textbf{S}^{ide}$, the grouping output $\textbf{G}$ becomes $\textbf{G}^{uk} \in \mathbb{R}^{\hbar_o \times \omega_o}$ where the element $g_i \rightarrow void$ if the corresponding semantic prediction $s_i \in [0,...,N-1]$. Afterward, $\textbf{S}^{ide}$ is merged with $\textbf{G}^{uk}$ to form the GOSS output $\textbf{G}_{oss} \in \mathbb{R}^{2 \times \hbar_o \times \omega_o}$. For $i$th pixel in $\textbf{G}_{oss}$, $\textbf{goss}_i = (s_i, void)$ if $s_i\in[0, 1, ..., N-1]$ or $\textbf{goss}_i = (N, g_i)$ if $s_i=N$. The prediction $\textbf{G}_{oss}$ can be viewed as a map which is composed of a set of several segments (see `GOSS prediction' in Figure \ref{figure_baseline}).

\subsection{GOSS Segmentor}
As shown in Figure \ref{figure_baseline}, the baseline model is extended to a new model that we call GOSS Segmentor (GST). Keeping the original five components of the baseline, we propose to equip the baseline with a confidence adjustment module and contrastive learning module.

\subsubsection{Confidence Adjustment}
For the N+1-model, it is hard to accommodate unknown classes of objects since the model is trained without seeing any examples from these classes. Instead of completely trusting the confidence prediction $\textbf{C}$, specific to the pixel identification of the N+1-model, we propose to modify $\textbf{C}$ using a confidence adjustment. Particularly, for the $i$th pixel, its confidence score after softmax, $\textbf{c}_i = [\textbf{c}_i^{kn}, c_i^{uk}] \in \mathbb{R}^{N+1}$, is re-scaled as $[\textbf{c}_i^{kn}, \beta^{uk} c_i^{uk}]$ where $\beta^{uk} \in (1, +\infty)$ is the scale coefficient of the confidence of the unknown class.

\subsubsection{Pixel Contrastive Learning}
In order to learn more discriminative representations and better identify\&cluster unknown inputs, inspired by \cite{wang2021exploring}, we adopt a pixel-wise contrastive learning algorithm where we contrast embeddings with embeddings with a different semantic label. For the embedding of the $i$th pixel $\textbf{e}_i \in \mathbb{R}^{cn}$ in the feature map $\textbf{E} \in \mathbb{R}^{cn \times \hbar_o \times \omega_o}$ where $cn$ is the channel number, the positive samples are pixel embedding $\textbf{e}_i^{+}$ with the same ground truth label to $\textbf{e}_i$ in the same image, while the negative samples are pixel embeddings $\textbf{e}_{i}^{-}$ having different ground truths. The pixel to pixel contrastive loss  \cite{wang2021exploring,sohn2016improved} is then defined as:
\begin{equation}
\resizebox{0.9\hsize}{!}{
$\ell_{pc, i} = \frac{1}{|\mathrm{P}_i|} \smashoperator{\sum_{i^+ \in \mathrm{P}_i}} -\log \frac{\exp({\textbf{e}_i \cdot \textbf{e}_{i}^{+}}/\tau)}{\exp({\textbf{e}_i \cdot \textbf{e}_{i}^{+}}/\tau)+ \smashoperator{\sum_{i^{-} \in \mathrm{N}_i}} \exp({\textbf{e}_i \cdot \textbf{e}_{i}^{-}}/\tau)}$
}
\end{equation}
\noindent 
where $\mathrm{P}_i$, $N_i$ are the set of positive and negative samples for pixel embedding $\textbf{e}_i$ and $\tau$ is the temperature parameter. We employ the semi-hard example sampling strategy from \cite{wang2021exploring} to construct the positive and negative samples sets. 

The pixel contrastive loss is merged with the classification loss $\ell^{cla}$ and the clustering loss $\ell^{clu}$ to give the total loss $\ell_{ws}$ of GST as: $\ell_{ws} = \alpha_{cla}\ell_{cla} + \alpha_{clu}\ell_ {clu} + \alpha_{pc}\ell_{pc}$ where $\alpha_{pc}$ is positive adjustment weights to $\ell_{pc}$. The purpose of a contrastive learning schema is to make the representations of pixels in the latent space from the same class closer, and further away from different classes. Many existing works have utilized similar metric learning techniques for segmentation tasks and gain better empirical results~\cite{harley2017segmentation,de2017semantic,Hwang_2019_ICCV,zhao2019region}. In GOSS, we apply contrastive learning to model training on known close-set data and hope the trained model can generate more representative embeddings for known and unknown pixels in open-set evaluation.

\section{Benchmark}
\label{section:dataset}
Most datasets for OSS, like StreetHazards \cite{hendrycks2019scaling} and Road Anomaly \cite{lis2019detecting}, present separate unknown objects in an image. Ensuring that objects of unknown classes appear together (are adjacent) in the natural image, in this work, using a proper ratio, we split the full set of labeling categories into known and unknown classes. We simulate the training and testing of GOSS using existing semantic segmentation datasets, i.e., COCO-Stuff \cite{caesar2018coco} and Cityscapes \cite{cordts2016cityscapes}. Note that grouping labels of unknown segments are derived from their initial ground-truth semantic labels before the split. Following \cite{wan2020super}, we convert the original semantic labeling of unknown areas to GS ground truths using connectivity labeling.

\subsection{COCO-Stuff-GOSS}
COCO-Stuff \cite{caesar2018coco} augments the popular COCO \cite{lin2014microsoft} dataset with stuff classes as well as dense-pixel annotations. It has a large-scale semantic multi-class setting containing both the `things' and `stuff' classes. On COCO-Stuff, around 94$\%$ of the pixels are labeled with one semantic category, and the remaining are `unlabeled' pixels. We customize COCO-Stuff, creating a new benchmark named COCO-Stuff-GOSS. We strictly divide existing specific classes of COCO-Stuff into known and unknown classes. Training and testing images are selected from `train2017' and `val2017', respectively. Those categories which have been defined as unknown categories will not be represented in the training examples. Every selected testing example is composed of objects from the set of known categories as well as the set of unknown categories (or from only unknown categories). The statistics of the benchmark on different splits are shown in Table~\ref{table_data_split}. 

\noindent \textbf{VOC Split.} The `VOC Split' is a common category split \cite{pinheiro2015learning,Hu_2018_CVPR,joseph2021towards} that provides 20 `thing' classes defined in PASCAL VOC \cite{everingham2010pascal} as `known thing' classes. The remaining 60 `thing' classes and 91 `stuff' classes are chosen as `unknown' classes.

\noindent \textbf{Manual Split.}
We divide COCO-Stuff categories according to how frequently each specific class appears. We count the number of occurrences of each class and calculate its ratio over the number of all training images. For example, in the `Manual-20/60' split, following that at least one and at most two classes are chosen from each sub-class, we choose 20 of the most popular `thing' classes and treat the remaining `thing' classes as unknown. Besides, all `stuff' classes are set as unknown classes.

\noindent \textbf{Random Split.} 
We also conduct experiments with a `Random Split', where all classes are randomly re-defined into known classes and unknown classes regardless of their super-class and sub-class. The data split of VOC-20/60 and Manual-20/60 do not include `stuff' categories as unknown classes, but Random-111/60 ensures that the known (or unknown) class includes specific classes from both the ‘thing’ and ‘stuff’ super-class. More details can be found in Table~\ref{table_data_split}. 

\subsection{Cityscapes-GOSS}
The Cityscapes \cite{cordts2016cityscapes} dataset consists of 5000 images (2975 train, 500 val, 1525 test) covering urban street scenes in driving scenarios. Dense pixel annotations of 19 classes are provided, where 8 of them are with instance-level segmentation, that is, 8 `thing' and 11 `stuff' classes.
As one goal of the proposed GOSS is to advance self-driving systems, we construct the Cityscapes-GOSS Benchmark. We divide the categories under the `manual split'. As opposed to the COCO-Stuff-GOSS Benchmark, all images, regardless of containing unknown categories or not, are kept. We consider pixels from unknown classes as `void' pixels. Table~\ref{table_data_split} presents more details.

\noindent \textbf{Manual Split.}
We present two versions of the Cityscapes-GOSS Benchmark. Following the split in \cite{Cen_2021_ICCV}, we build the first version, `Manual-16/3', which includes `car', `truck', and `bus' as the `unknown thing'. Based on the first version, we additionally make `building', `traffic sign', and `vegetation' as `unknown stuff', to produce a more challenging version, `Manual-13/6'.

\begin{table*}
\begin{center}
\resizebox{1.\textwidth}{!}{
\begin{tabular}{l|c|cc|cc|cc|cc}
\hline
\multirow{2}*{\textbf{Dataset}} & \multirow{2}*{\textbf{Data Split}}  & \multirow{2}*{\textbf{Known}} & \multirow{2}*{\textbf{Unknown}} 
& \multirow{2}*{\makecell[c]{\textbf{Known} \\ \textbf{(thing)}}} & \multirow{2}*{\makecell[c]{\textbf{Unknown} \\ \textbf{(thing)}}}
& \multirow{2}*{\makecell[c]{\textbf{Known} \\ \textbf{(stuff)}}} & \multirow{2}*{\makecell[c]{\textbf{Unknown} \\ \textbf{(stuff)}}}  
& \multirow{2}*{\makecell[c]{\textbf{Train} \\ \textbf{images}}} & \multirow{2}*{\makecell[c]{\textbf{Test} \\ \textbf{images}}} \\
&&&&&&&&& \\
\hline
COCO-Stuff \cite{caesar2018coco}    & -                       &171  &0   &80  &0   &91  &0   &118287  &5000 \\ \hline
\multirow{3}*{COCO-Stuff-GOSS}          & VOC-20/60      &111  &60  &20  &60  &91  &0   &21711   &4080 \\
                                        & Manual-20/60   &111  &60  &20  &60  &91  &0   &17293   &4264 \\
                                        & Random-111/60  &111  &60  &51  &29  &60  &31  &18707   &4156 \\ 
\hline

Cityscapes \cite{cordts2016cityscapes}  & -  &19  &0  &8  &0  &11  &0   &2975  &500 \\ \hline
\multirow{2}*{Cityscapes-GOSS}     & Manual-16/3    &16  &3  &5  &3  &11  &0   &2975   &500 \\
                                   & Manual-13/6    &13  &6  &5  &3  &8   &3   &2975   &500 \\ \hline
\end{tabular}
}
\end{center}
\caption{Details of different splits of the COCO-Stuff-GOSS/Cityscapes-GOSS benchmarks. The numbers in the table indicate for each data split, how many known (or unknown) classes are selected and how many training (or testing) images are kept. We also provide details of the original COCO-Stuff~\cite{caesar2018coco} and Cityscapes~\cite{cordts2016cityscapes} for comparison.} \label{table_data_split}
\end{table*}
\section{Experiments}
Experimental results are presented in this section to demonstrate the rationality and effectiveness of GOSS. We perform our novel task on COCO-Stuff-GOSS and Cityscapes-GOSS using the baseline and proposed GST. The performance is mainly measured via the metric $\mathrm{GQ}$.

\subsection{Implementation}
For all models, ResNet-50 \cite{he2016deep} pre-tained on ImageNet \cite{deng2009imagenet} is utilized as the encoder backbone. All models are trained for 60K/40K iterations with a batch size of 10/2 on COCO-Stuff-GOSS/Cityscapes-GOSS. The initial learning rate is set to $5\mathrm{e}{-5}$. The weights $\alpha_{cla}$, $\alpha_{clu}$ and $\alpha_{pc}$ are 1.0, $1\mathrm{e}{-4}$ and $1\mathrm{e}{-2}$. The thresholds in the identification module and the scale $\beta^{uk}$ (for +CA) are set to 0.5 and 5.0, respectively. Our models are implemented in PyTorch~\cite{paszke2019pytorch}. We note that the N+1-model cannot be used for the Cityscape-GOSS dataset. As we consider labels of unknown classes to be `void' instead of directly filtering out the image, the entropy of void pixels is not allowed to be added into the loss.

\subsection{Results}
\label{results}
The results of GOSS on COCO-Stuff-GOSS and Cityscapes-GOSS using various identification methods are reported in Table \ref{table_ws} and \ref{table_ws_city}, respectively. In addition to GOSS quality, we also provide metrics for OSS (AUROC and AURP) and GS (mIoU and $\mathrm{GQ}^{clu}$) tasks to show that the models perform reasonably on these relevant older tasks. 

For COCO-Stuff-GOSS in Table \ref{table_ws}, GST becomes the best performing model. For instance, on the `Manual-20/60 split', GST attains 9.15$\%$ $\mathrm{GQ}$, outperforming the N-model+MSP by a healthy margin of nearly 1.45$\%$. Compared to other baselines, the pixel contrastive learning module assists GST to better discriminate between the known pixels and the unknown pixels in most cases (See `OSS Metric' in Table \ref{table_ws}). Moreover, it boosts the clustering accuracy in GS. One of the baseline models, N-model+MaxLogit with using threshold sacrifices much of $\mathrm{GQ}^{kn}$, but it achieves a high $\mathrm{GQ}^{uk}$. As expected, MaxLogit identifies more unknown areas, however, it does not simultaneously maintain the pixel classification accuracy \cite{jung2021standardized}. For Cityscapes-GOSS in Table \ref{table_ws_city}, we find a similar performance ranking to COCO-Stuff-GOSS in Table \ref{table_ws}. 

\begin{table*}
\begin{center}
\resizebox{0.98\textwidth}{!}{
\begin{tabular}{l|l|c|cc|cc|ccc}
\hline
&& &\multicolumn{2}{c|}{\textbf{OSS Metric}} &\multicolumn{2}{c|}{\textbf{GS Metric}} &\multicolumn{3}{c}{\textbf{GOSS Metric}} \\ \hline

\textbf{Data Split} &\makecell[c]{\textbf{Identification}\\\textbf{Method}}  &\makecell[c]{\textbf{Clustering} \\ \textbf{Method}} &\textbf{AUROC} &\textbf{AUPR} &\textbf{mIoU} &$\textbf{GQ}^{\textbf{clu}}$ &$\textbf{GQ}^{\textbf{kn}}$ &$\textbf{GQ}^{\textbf{uk}}$ &$\textbf{GQ}$ \\ \hline

\multirow{5}*{VOC-20/60}     &N-model+MSP \cite{hendrycks2016baseline}      &\multirow{5}*{super-BPD} &76.5 &26.4 &12.7 &27.3 &14.4 &3.0 &8.70 \\
                             &N-model+Maxlogit \cite{hendrycks2019scaling}  &                         &71.0 &22.0 &12.7 &27.3 &3.6  &4.3 &3.93 \\           
                             &N-model+DML \cite{Cen_2021_ICCV}              &                         &71.1 &20.9 &12.5 &26.6 &1.6  &4.3 &2.93 \\ 
                             &N+1-model \cite{devries2018learning}          &                         &75.7 &25.7 &12.4 &25.9 &14.3 &3.3 &8.83 \\
                             &GST(Ours)                                     &                         &77.0 &27.0 &13.6 &28.7 &15.4 &4.9 &\textbf{9.84} \\ \hline

\multirow{5}*{Manual-20/60}  &N-model+MSP \cite{hendrycks2016baseline}      &\multirow{5}*{super-BPD} &74.7 &27.8 &12.8 &26.1 &13.1 &2.3 &7.70 \\
                             &N-model+Maxlogit \cite{hendrycks2019scaling}  &                         &71.4 &24.1 &12.8 &26.1 &3.7 &4.5 &4.13 \\           
                             &N-model+DML \cite{Cen_2021_ICCV}              &                         &75.9 &28.5 &12.8 &27.3 &13.3 &2.3 &7.81 \\ 
                             &N+1-model \cite{devries2018learning}          &                         &76.2 &29.5 &12.9 &27.4 &13.1 &2.2 &7.62 \\
                             &GST(Ours)                                     &                         &77.3 &30.7 &13.8 &28.3 &14.3 &3.9 &\textbf{9.15} \\ \hline

\multirow{5}*{Random-111/60} &N-model+MSP \cite{hendrycks2016baseline}      &\multirow{5}*{super-BPD} &77.5 &43.1 &12.6 &27.9 &16.6 &4.1 &10.36 \\
                             &N-model+Maxlogit \cite{hendrycks2019scaling}  &                         &73.9 &41.2 &12.6 &27.9 &3.7 &8.7 &6.22 \\           
                             &N-model+DML \cite{Cen_2021_ICCV}              &                         &78.6 &44.4 &12.7 &28.5 &17.0 &4.3 &10.67\\ 
                             &N+1-model \cite{devries2018learning}          &                         &77.9 &44.9 &12.6 &27.1 &16.1 &4.2 &10.16 \\
                             &GST(Ours)                                     &                         &79.8 &47.0 &13.7 &29.0 &17.5 &4.2 &\textbf{10.87}\\ \hline
\end{tabular}
}
\end{center}
\caption{GOSS results of GST (N+1-model+CA+CL) on COCO-Stuff-GOSS under three splits. `CA' is the confidence adjustment and `CL' is the cross-pixel contrastive learning. The best results of $\mathrm{GQ}$ are in \textbf{bold}.}\label{table_ws}
\end{table*}

\begin{table*}
\begin{center}
\resizebox{0.90\textwidth}{!}{
\begin{tabular}{l|l|c|cc|cc|ccc}
\hline
&& &\multicolumn{2}{c|}{\textbf{OSS Metric}} &\multicolumn{2}{c|}{\textbf{GS Metric}} &\multicolumn{3}{c}{\textbf{GOSS Metric}} \\ \hline

\textbf{Data Split} &\makecell[c]{\textbf{Identification} \\ \textbf{Method}}  &\makecell[c]{\textbf{Clustering} \\ \textbf{Method}} &\textbf{AUROC} &\textbf{AUPR} &\textbf{mIoU} &$\textbf{GQ}^{\textbf{clu}}$ &$\textbf{GQ}^{\textbf{kn}}$ &$\textbf{GQ}^{\textbf{uk}}$ &$\textbf{GQ}$ \\ \hline

\multirow{4}*{Manual-16/3}   &N-model+MSP \cite{hendrycks2016baseline}      &\multirow{4}*{super-BPD} &82.2 &59.3 &7.2 &3.8 &12.1 &1.1 &6.60 \\
                             &N-model+Maxlogit \cite{hendrycks2019scaling}  &                         &79.8 &57.4 &7.2 &3.8 &6.2 &1.7 &3.92 \\           
                             &N-model+DML \cite{Cen_2021_ICCV}              &                         &83.1 &61.2 &6.9 &3.0 &12.3 &1.1 &6.74 \\ 
                             &GST(Ours)                                     &                         &83.8 &62.6 &7.0 &3.5 &13.4 &1.1 &\textbf{7.31} \\ \hline
                             
\end{tabular}
}
\end{center}
\caption{GOSS results of GST (N-model+CL) on Cityscapes-GOSS under `Manual-16/3' split. `CL' is the cross-pixel contrastive learning. The best results of $\mathrm{GQ}$ are in \textbf{bold}. The results of `Manual-13/6' split are provided in the supplementary material. 
}\label{table_ws_city}
\end{table*}

\begin{figure*}
\centering
	\includegraphics[width=1.0\linewidth]{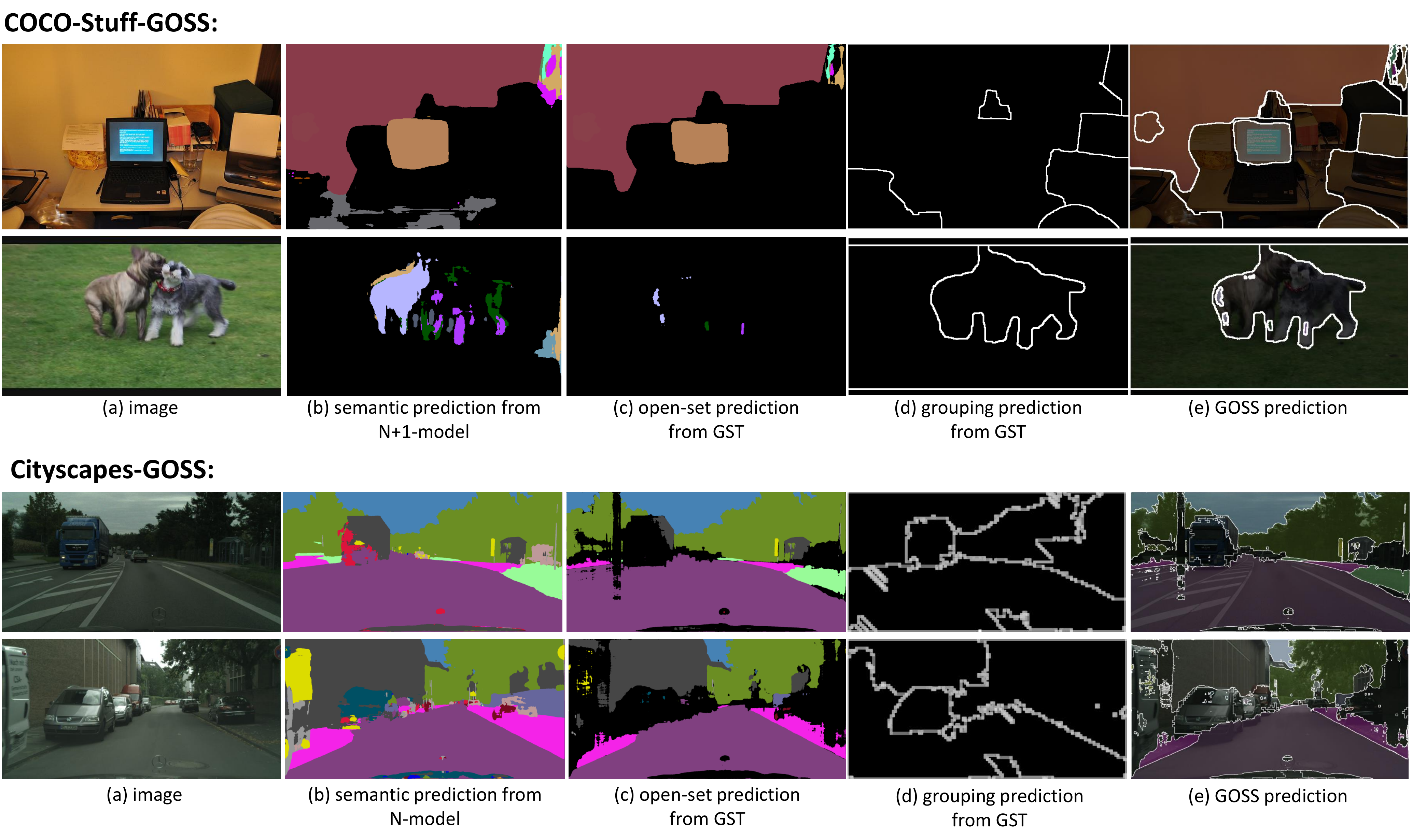}
 	\caption{Visualized segmentation results from GST (N+1-model+CA+CL/N-model+CL). The GOSS prediction (e) merges the OSS prediction (c) and the grouping prediction (d). Hence, within GOSS prediction (e), `objects' inside identified unknown regions can be segmented out. For example, unknown objects, `paper'/`printer' in the 1st-row and, `dog'/`grass' in the 2nd-row are correctly outlined even though their classes are not known.}
 	\label{figure_examples}
\end{figure*}

Several examples from two built benchmarks are visualized in Figure \ref{figure_examples} for better illustrating the GOSS settings. Taking one example from Figure \ref{figure_examples} (2nd-row figure), GOSS accurately segments out `unknown dogs' from `unknown grass' (see Figure \ref{figure_examples} (e)). Compared to the prediction of OSS in Figure \ref{figure_examples} (c), GOSS can provide richer information for intelligent agents to make decisions. In terms of GST model, it can be observed from Figure \ref{figure_examples} (1st figure and 2-nd figure) that the confidence adjustment module of GST effectively helps the N+1-model to detect more unknown regions (see Figure \ref{figure_examples} (b) and (c)).

\subsection{Analysis}

\noindent \textbf{Training Strategy.} For all models in Table \ref{table_ws} and \ref{table_ws_city}, the pixel classification branch and the pixel clustering branch are trained in a unified single architecture. Here, we study a different training strategy, `Separate', where two branches are trained separately, and then their outputs are merged. We find that the performances of `Separate' and our `Single' network are close. We finally choose `Single' since it is fast, light, and easy to implement.

\begin{table}
\begin{center}
\resizebox{0.48\textwidth}{!}{
\begin{tabular}{l|l|c|c|ccc}
\hline
&&&  &\multicolumn{3}{c}{\textbf{GOSS Metric}} \\ \hline

\textbf{Data Split} &\makecell[c]{\textbf{Identification} \\ \textbf{Method}}  &\makecell[c]{\textbf{Clustering} \\ \textbf{Method}} &\textbf{Strategy} &$\textbf{GQ}^{\textbf{kn}}$ &$\textbf{GQ}^{\textbf{uk}}$ &$\textbf{GQ}$ \\ \hline

\multirow{4}*{Random-111/60} &N-model+MSP \cite{hendrycks2016baseline}      &\multirow{4}*{super-BPD} &\multirow{2}*{Single}   &16.6 &4.1 &10.36 \\
                             &N-model+Maxlogit \cite{hendrycks2019scaling}  &                         &                         &3.7  &8.7 &6.22 \\ \cline{2-2} \cline{4-7}
                             &N-model+MSP \cite{hendrycks2016baseline}      &                         &\multirow{2}*{Separate} &16.4 &4.2 &10.17 \\
                             &N-model+Maxlogit \cite{hendrycks2019scaling}  &                         &                         &3.4 &8.9 &6.14 \\ \hline
\end{tabular}
}
\end{center}
\caption{Ablation study on COCO-Stuff-GOSS under `Random-111/60' split: training strategy. Two strategies are compared.}\label{table_ws_ablation2}
\end{table}

\noindent \textbf{Clustering Method.}
We train the pixel clustering branch for grouping unknown areas in an unsupervised manner applying differentiable feature clustering (DFC) loss \cite{kim2020unsupervised_simple}. DFC in an unsupervised setting has a basic clustering performance, but it is worse than Super-BPD in a supervised setting.

\begin{table}
\begin{center}
\resizebox{0.48\textwidth}{!}{
\begin{tabular}{l|l|c|cc|ccc}
\hline
&&  &\multicolumn{2}{c|}{\textbf{GS Metric}} &\multicolumn{3}{c}{\textbf{GOSS Metric}} \\ \hline

\textbf{Data Split} &\makecell[c]{\textbf{Identification} \\ \textbf{Method}}  &\makecell[c]{\textbf{Clustering} \\ \textbf{Method}} &\textbf{mIoU} &$\textbf{GQ}^{\textbf{clu}}$ &$\textbf{GQ}^{\textbf{kn}}$ &$\textbf{GQ}^{\textbf{uk}}$ &$\textbf{GQ}$ \\ \hline

\multirow{4}*{Random-111/60} &N-model+MSP \cite{hendrycks2016baseline}      &\multirow{2}*{super-BPD} &12.6 &27.9 &16.6 &4.1 &10.36 \\
                             &N-model+Maxlogit \cite{hendrycks2019scaling}  &                         &12.6 &27.9 &3.7  &8.7 &6.22 \\ \cline{2-8}
                             &N-model+MSP \cite{hendrycks2016baseline}      &\multirow{2}*{DFC}       &4.8 &4.7 &16.8 &1.8 &9.31 \\
                             &N-model+Maxlogit \cite{hendrycks2019scaling}  &                         &4.8 &4.7 &3.7 &3.6 &3.69 \\ \hline
\end{tabular}
}
\end{center}
\caption{Ablation study on COCO-Stuff-GOSS under `Random-111/60' split: clustering method. Two clustering models of GST, super-BPD in a supervised setting and DFC \cite{kim2020unsupervised_simple} in an unsupervised setting are compared.}\label{table_ws_ablation1}
\end{table}

\noindent \textbf{Challenging Task.}
The results in Table \ref{table_ws} and \ref{table_ws_city} verify that GOSS is a very challenging task, despite our baseline framework relying on strong backbones and a reasonable architecture. The first main reason is that it is non-trivial to perform accurate pixel identification under the open-set setting. Furthermore, the clustering branch suffers a performance drop when the model encounters the unfamiliar appearances of objects from unknown categories at test-time. There is significant room for future improvement on the task of GOSS.

\section{Conclusion}
The novel setting referred to as GOSS is introduced in this paper. Our goal is to build upon the well-defined OSS towards generating more comprehensive predictions. The task is to semantically classify pixels as one of known classes or an unknown class as well as cluster the detected unknown pixels. With more extracted information inside the unknown region, GOSS is beneficial for intelligent agents in their decision making process. Specific to the new setting, a metric, two benchmarks, and a corresponding baseline model are presented. In future works, the concept of GOSS can be further extended to include instance segmentation, image co-segmentation, video segmentation, point cloud segmentation, etc. We hope that this work may provide a new alternative towards a more comprehensive pixel-level scene understanding.

{\small
\bibliographystyle{ieee_fullname}
\bibliography{egbib}
}

\end{document}